# An Evolutionary Approach to Drug-Design Using Quantam Binary Particle Swarm Optimization Algorithm

Avishek Ghosh, Arnab Ghosh, Arkabandhu Chowdhury
Dept. of Electronics & Telecommunication Engineering
Jadavpur University
Kolkata 700032, India
avishek.ghosh38@gmail.com,arnabju90@gmail.com,
arjia_2005@yahoo.com

Jubin Hazra
Dept. of Electronics & Communication Engineering
Institute of Engineering & Management
Kolkata 700091, India
jubinhazra.iem@gmail.com

*Abstract*— **The present work provides a new approach to evolve ligand structures which represent possible drug to be docked to the active site of the target protein. The structure is represented as a tree where each non-empty node represents a functional group. It is assumed that the active site configuration of the target protein is known with position of the essential residues. In this paper the interaction energy of the ligands with the protein target is minimized. Moreover, the size of the tree is difficult to obtain and it will be different for different active sites. To overcome the difficulty, a variable tree size configuration is used for designing ligands. The optimization is done using a quantum discrete PSO. The result using fixed length and variable length configuration are compared.**

## I. INTRODUCTION

A strategy in drug design is to find compounds that bind to protein targets that constitute active sites which sustain viral proliferation. In many literatures [7],[8],[9],[10] interactions between protein molecules (protein-protein docking) are reported. This work deals with the interaction of protein with small drug molecule called ligands. The challenge is to predict accurately structures of the compounds (ligands) when the active site configuration of the protein is known [1]. The literature addresses the challenge using a novel quantum discrete PSO algorithm that uses two different populations to generate offspring. It is found that the algorithm gives better candidate solution than traditional Binary PSO Algorithm.

Evolutionary computation is used to place functional groups in appropriate leaves of the tree structured ligand. The objective is to minimize the interaction energy between the target protein and the evolved ligand, thus leading to the most stable solution. In [1] a fixed tree structure of the ligand is assumed. However it is difficult to get a prior knowledge of the structure and for a given geometry, no unique solution is the best solution. So, variable length structure is used in the paper. Depending upon the geometry of the active site, a ligand can have a maximum or a minimum length (denoted by $l_{max}$ and $l_{min}$). The length of the ligand lies in between these two values.

## II. ALGORITHM DESCRIPTION

### A. Binary PSO Algorithm

Discrete Binary Particle swarm optimizer (BPSO) is introduced by Kennedy and Eberhart in 1995 [2]. In the binary version of PSO each particle is represented by a string of zeroes and ones. In case of binary PSO, the particle's personal best and global best is updated as in continuous version [3]. The major difference between binary PSO with continuous version is in the way velocity is defined. Velocity is updated in similar way as in continuous version but the velocity is mapped from real value to the range [0,1]. Sigmoid function is used as the normalized function to do the mapping. Velocity mapping for the $i^{th}$ particle for $d^{th}$ dimension can be defined as:

$$sig(v_{i,d}(t)) = \frac{1}{1+e^{-v_{i,d}(t)}} \quad (1)$$

Velocity is updated as in equation (2).

$$v_{i,d}(t+1) = \omega.v_{i,d}(t) + \varphi_1.rand1_{i,d}(0,1).(p^l_{i,d} - x_{i,d}(t)) + \varphi_2.rand2_{i,d}(0,1).(p^g_d - x_{i,d}(t)) \quad (2)$$

where $w$ represents the inertia factor and $\varphi_1$, $\varphi_2$ two positive integer to give the weightage for local best and global best of particles.

New position of particle can be found using the following equation.

$$x_{i,d}(t) = \begin{cases} 1 & \text{if } rand_{i,d}(0,1) < sig(v_{i,d}(t)) \\ 0 & \text{otherwise} \end{cases} \quad (3)$$

So, the velocity $v_{i,d}(t)$ component become the probability parameter for $x_{i,d}(t)$ to be set (1) or clear (0).

## B. Quantum Discrete PSO

The main problem lies in the binary version of the PSO proposed by Kennedy and Eberhart lies in the parameter selection [4]. Some parameter effect is found to be opposite to that of the real valued PSO. For binary PSO small value of $v_{max}$ promotes exploration. Choosing the inertia factor is the major difficulty in case of binary PSO. For binary PSO, values of $w$ less than 1 prevent convergence. For values of $-1 < w < 1$, velocity component becomes 0 over time. If w is greater than 1 velocity increases over time and all bits become 1. If inertia factor is not chosen properly the PSO particles or dynamics may not converge. A new variant of binary version of PSO is proposed by Yang in 2004 [5]. In this version two different populations are maintained.

A population of quantum particle vectors:

$$\mathbf{V} = [\vec{V}_1, \vec{V}_2, ..., \vec{V}_M], \quad (\vec{V}_i = [v_i^1, v_i^2, ..., v_i^N])$$

where $0 \leq v_i^j \leq 1$ $(i = 1,2,...,M; j = 1,2,...,N)$  (4)

A population of discrete particle vectors:

$$\mathbf{X} = [\vec{X}_1, \vec{X}_2, ..., \vec{X}_M], \quad (\vec{X}_i = [x_i^1, x_i^2, ..., x_i^N])$$

where $x_i^j \in \{0,1\}$, $(i = 1,2,...,M; j = 1,2,...,N)$  (5)

M represents the number of particle and N is the dimension of each particle.

Position update is done by flipping rule:

$$\begin{aligned} x_i^j &= 0, \quad \text{if } rand(0,1) \geq v_i^j; \\ &= 1, \quad \text{otherwise.} \end{aligned} \quad (6)$$

Quantum populations are updated as follow:

$$v_{local\_best} = \alpha \times x_{local\_best} + \beta \times (1 - x_{local\_best}),$$
$$v_{global\_best} = \alpha \times x_{global\_best} + \beta \times (1 - x_{global\_best}),$$
$$v_{new} = \omega \times v_{old} + c_1 \times v_{local\_best} + c_2 \times v_{global\_best}$$

where $\alpha + \beta = 1$, $0 < \alpha, \beta < 1$

and $\omega + c_1 + c_2 = 1$, $0 < \omega, c_1, c_2 < 1$  (7)

where $w$ is the inertia factor and $c_1$, $c_2$ are the acceleration coefficients.

The problem regarding non-convergence of binary PSO can be overcome with this algorithm.

## III. EVOLVING LIGAND MOLECULE

Protein with known active site configuration is used for evolving ligand structures. Our specific target is the known antiviral binding site of the Human Rhinovirus strain 14. This active site is known as the VP1 barrel for its resemblance with a barrel. The molecule which can easily be fit in the structure having minimum interaction energy will be the evolved drug (ligand). For simplification, a 2-dimensional structure is chosen [1].

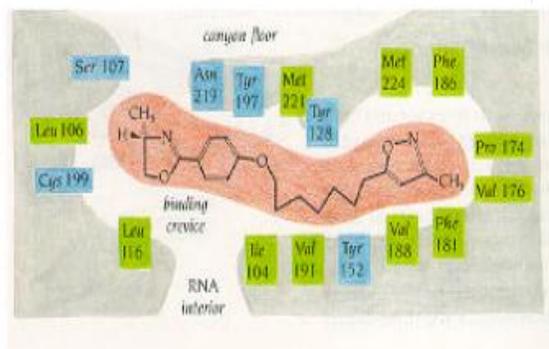

Figure 1.    Active site of Human Rhinovirus strain 14

Figure 1 illustrates the binding site of Human Rhinovirus strain 14 and a typical structure of ligand is illustrated. For designing the ligand, the co-ordinates of the residues of the protein must be known. A ligand molecule is assumed to have a tree like structure on both sides of a fixed pharmacophore illustrated in figure 3. The structure has a left hand side known as left tree and a right hand side called right tree. For this configuration the right hand side contains 10 leaves and the left hand side contains 7 leaves. Each functional group is represented by a 3 bit binary code. Absence of a group is denoted by binary code 000 (0-group). Each leaf represents a functional group among the 7 functional groups listed in figure 2 along with their bond length projection on the x axis (table 1). It is to be noted that when group 0 occupies a position, the length of the tree is reduced enabling a variable length tree structure. For fixed length tree, group 0 is absent. The job is to find appropriate functional groups for the leaves such that the protein ligand docking energy (interaction energy) is minimized.

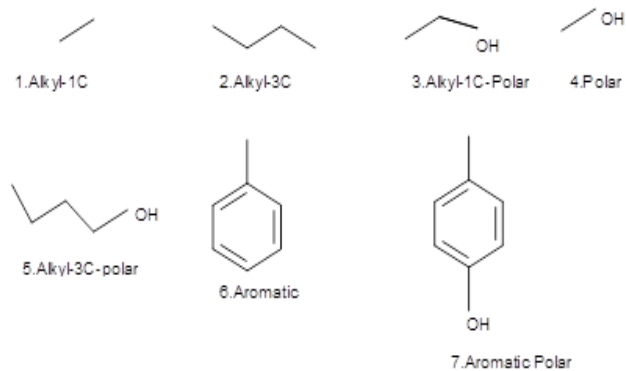

Figure 2.    Functional Groups

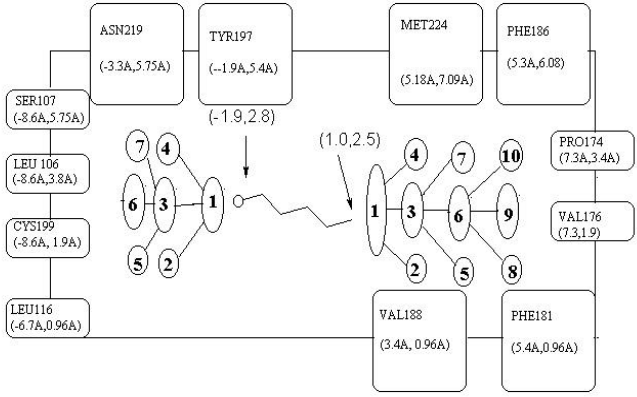

Figure 3. Active site co-ordinates of Human Rhinovirus strain 14

TABLE I. BOND LENGTH AND BINARY CODE OF FUNCTIONAL GROUPS

| Functional group | Bond length along x axis (Å) | Binary Code |
|---|---|---|
| NUL (no group) | - | 000 |
| Alkyl-1C | 0.65 | 001 |
| Alkyl-3C | 1.75 | 010 |
| Alkyl-1C-Polar | 1.1 | 011 |
| Alkyl-3C-Polar | 2.2 | 100 |
| Polar | 0.01 | 101 |
| Aromatic | 1.9 | 110 |
| Aromatic-Polar | 2.7 | 111 |

After computing the co-ordinates of each functional group, the Eucledian distances between the group and all the residues of the active site are computed to find the Van der Waals energy.

## IV. VARIABLE LENGTH ALGORITHM

The size of representing tree is made variable. The length of the chromosome can vary between certain range denoted by $(l_{max}, l_{min})$, where these two are defined as:

$$l_{max} = \frac{\text{length of the major axis}}{\text{maximum bond length}}$$

$$l_{min} = \frac{\text{length of the major axis}}{\text{minimum bond length}} \quad (8)$$

$l_{min}$ is calculated 7 for right tree and 2 for left tree. Binary code 0 (000) is used to represent the absence of a functional group. One extra function named "Correct" is used to check correctness of the newly formed chromosome. Some constrains should be maintained. The 1st, 3rd and 6th position of right tree and 1st and 3rd position of left tree cannot contain any type of polar group unless they are the terminals of the tree. If any of the positions among 8th, 9th, 10th of the right tree contain any group then 6th position cannot contain 0. Similarly for left tree if any of the positions among 5th, 6th, and 7th contain any group then 3rd position cannot contain 0.

## V. FITNESS EVALUATIONS

The computation of fitness was based on the interaction energy of the residues with the closest functional group and the chemical properties of these pairs. The distance between residues and functional groups should not be more than 2.7Å and less than 0.7Å. If the functional group and the closest residue are of different polarity, a penalty is imposed. The interaction energy of the ligand with the protein is the sum of Van der Waals potential energy between the groups and the amino acid residues present in the active site of the target protein. The Van der Waals potential energy is computed as:

$$V(r) = \left[\left(\frac{c_n}{r^6}\right) - \left(\frac{c_m}{r^{12}}\right)\right] \quad (9)$$

Where $n$ and $m$ are integers and $c_n$ and $c_m$ are constants [6]. Finally fitness $F$ is computed as

$$F = \frac{k}{E} \quad (10)$$

Where $k$ is a constant (typically 100) and E is the total interaction energy in Kcal/mol. Therefore, maximizing fitness leads to minimizing interaction energy.

## VI. RESULTS

The initial population size is taken to be 40 and the algorithm is run for 100 generations. The interaction energy using fixed length Quantum Binary PSO and variable length Quantum Binary PSO is tabulated in table 2. The configuration of the evolved ligand for both fixed length and variable length algorithm is drawn in the figure 5 and figure 6. It is found that variable length structures lead to more stable and fit candidate solution than the fixed length structure. The fitness of best candidate solution is plotted against generation ($k$ is taken 100) in figure 4.

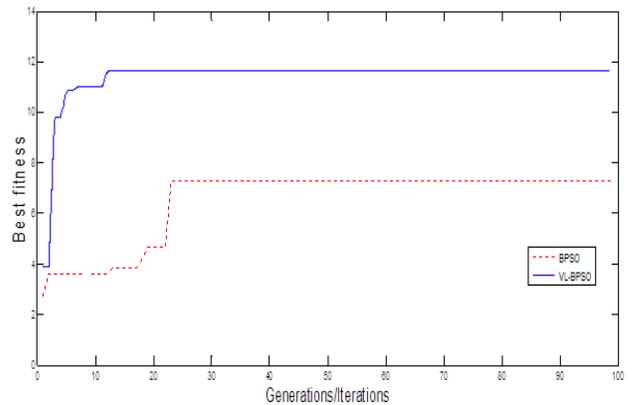

Figure 4. Computation of fitness over generations

TABLE II. INTERACTION ENERGY VALUES CORRESPONDING TO HUMAN RHINOVIRUS STRAIN 14

| Algorithm | Interaction Energy (Kcal/Mol) |
|---|---|
| Variable length BPSO | 8.5707 |
| Fixed length BPSO | 12.4489 |

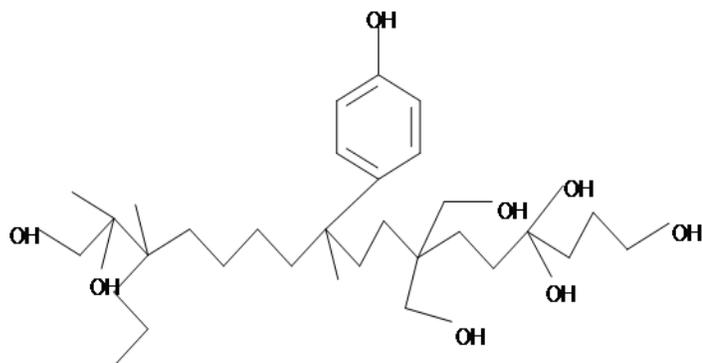

Figure 5. Fixed length structure

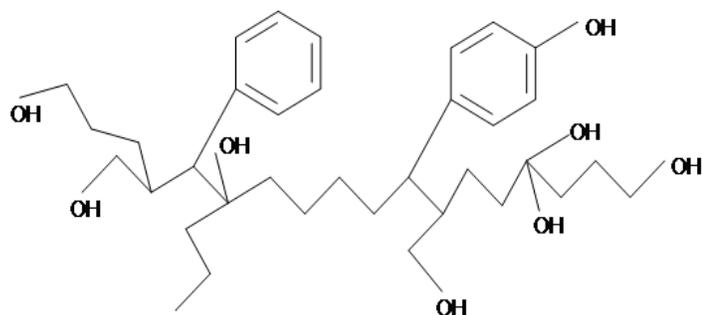

Figure 6. Variable length structure

## VII. CONCLUSIONS

We can conclude that Quantum binary Particle Swarm optimization algorithm gives better result for ligand design problem when the ligand length is considered as variable. The paper uses a 2-dimensional approach which is quite unrealistic. A 3-dimensional approach using more complex tree structure will be our future research goal. Although protein-ligand interaction energy is minimized in the current paper, it may be required in some cases to minimize ligand energy to find stable solution. A multi-objective approach to minimize protein ligand interaction energy as well as ligand energy will be our future endeavor.

REFERENCES

[1] Goh G, Foster JA, "Evolving Molecules for Drug Design Using Genetic Algorithm", Proc. Int. Conf. on Genetic & Evol. Computing, Morgan Kaufmann, 27 – 33, 2000.
[2] Kennedy, J.; Eberhart, R.C. "A discrete binary version of the particle swarm algorithm", IEEE International Conference on Systems, Man, and Cybernetics, 1997.
[3] J. Kennedy, "Small worlds and mega-minds: effects of neighborhood topology on particle swarm performance". Proc. of IEEE Congress on Evolutionary Computation (CEC 1999), Piscataway, NJ. pp. 1931-1938, 1999.
[4] A. P. Engelbrecht, "Fundamentals of Computational Swarm Intelligence", Wiley, 2005.
[5] Shuyuan Yang Min Wang Licheng jiao, "A quantum particle swarm optimization", Evolutionary Computation, 2004. CEC2004.
[6] Leach AR, "Molecular Modeling Principles and Applications", Pearson, Prentice Hall, 2001.
[7] Luis Rueda, Sridip Banerjee, Md. Mominul Aziz, Mohammad Raza, "Protein-protein Interaction Prediction using Desolvation Energies and Interface Properties", Proceedings of IEEE International Conference on Bioinformatics and Biomedicine, pp: 17-22.
[8] Nazar Zaki, "Prediction of Protein–Protein Interactions Using Pairwise Alignment and Inter-Domain Linker" Region,Engineering Letter, 16:4, EL_16_4_07.
[9] Piyali Chatterjee, Subhadip Basu, Mahantapas Kundu, Mita Nasipuri and Dariusz Plewczynski "PPI_SVM: Prediction of Protein-protein Interactions using Machine Learning, Domain-domain Affinities and Frequency Tables", Cellular & Molecular Biology Letters, volume 16 (2011) pp 264-278.
[10] Jos´e A. Reyes and David Gilbert, "Prediction of protein-protein interactions using one-class classification methods and integrating diverse biological data", Journal of Integrative Bioinformatics 4(3):77, 2007